\pdfoutput=1

\documentclass[11pt]{article}

\usepackage{acl}

\usepackage{times}
\usepackage{latexsym}

\usepackage[T1]{fontenc}

\usepackage[utf8]{inputenc}

\usepackage{times}
\usepackage{latexsym}

\usepackage{booktabs}
\usepackage{multirow}
\usepackage{bm}
\usepackage{amsmath}
\usepackage{graphicx}
\usepackage{CJKutf8}
\usepackage{xcolor}
\usepackage{xpinyin}
\usepackage{pifont}
\usepackage{amssymb}

\usepackage{microtype}

\usepackage{url}

\usepackage{footnote}
\makesavenoteenv{table}

\title{Seeking Patterns, Not just Memorizing Procedures: \\
    Contrastive Learning for Solving Math Word Problems}

\author{Zhongli Li$^1$\thanks{\; Zhongli Li and Wenxuan Zhang contributed equally. Qingyu Zhou is the corresponding author.},~~Wenxuan Zhang$^{2*}$\thanks{\; Contribution done during internship at Tencent Cloud Xiaowei.},~~Chao Yan$^{2\dagger}$,~~Qingyu Zhou$^{1}$,\\\textbf{Chao Li}$^1$,~~\textbf{Hongzhi Liu}$^2$,~~\textbf{Yunbo Cao}$^1$
 \\
  $^1$Tencent Cloud Xiaowei \\
  $^2$Peking University \\
   \texttt{\{neutrali,qingyuzhou,diegoli,yunbocao\}@tencent.com} \\
   \texttt{\{zwx980624@stu,cyan@stu,liuhz@ss\}.pku.edu.cn}
}

\date{}

\begin{document}
\maketitle
\begin{abstract}

Math Word Problem~(MWP) solving needs to discover the quantitative relationships over natural language narratives.
Recent work shows that existing models memorize procedures from context and rely on shallow heuristics to solve MWPs.
In this paper, we look at this issue and argue that the cause is a lack of overall understanding of MWP patterns.
We first investigate how a neural network understands patterns only from semantics, and observe that, if the prototype equations like $n_1+n_2$ are the same, most problems get closer representations and those representations apart from them or close to other prototypes tend to produce wrong solutions.
Inspired by it, we propose a contrastive learning approach, where the neural network perceives the divergence of patterns. 
We collect contrastive examples by converting the prototype equation into a tree and seeking similar tree structures. The solving model is trained with an auxiliary objective on the collected examples, resulting in the representations of problems with similar prototypes being pulled closer.
We conduct experiments\footnote{The code is available at \url{https://github.com/zwx980624/mwp-cl}.} on the Chinese dataset Math23k and the English dataset MathQA. Our method greatly improves the performance in monolingual and multilingual settings. 




\end{abstract}

\section{Introduction}

\begin{figure}[t]
    \centering
      \includegraphics[width=7.5cm]{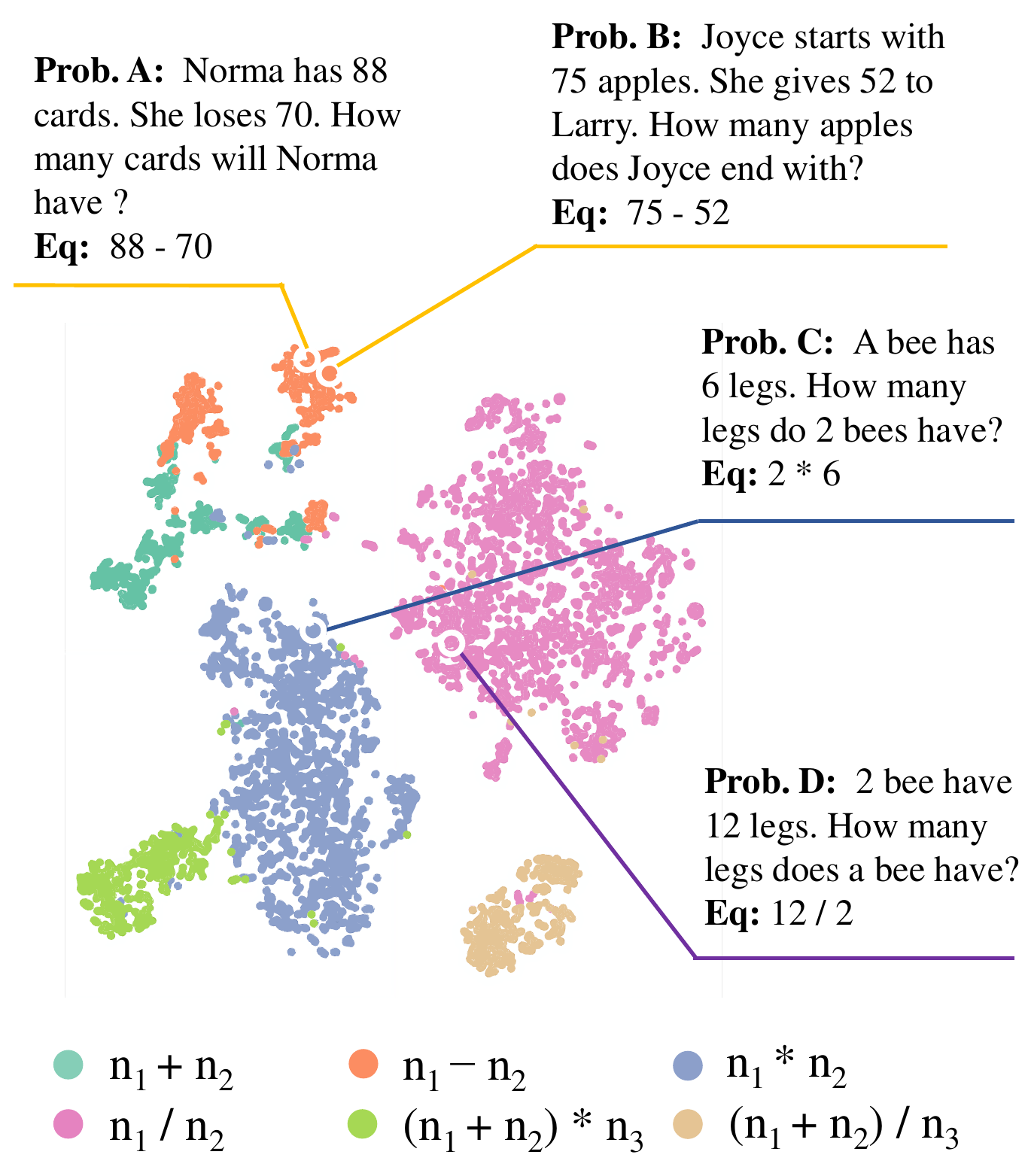}
    \caption{The visualization of the problem representations by T-SNE. "Prob." and "Eq" are short for the math word problem and its solution equation. The problem A and B are in the same prototype equation "$n_1-n_2$". The problem C and D are semantically similar.}
    \label{fig:intro-tsne}
\end{figure}

A Math Word Problem~(MWP) is described as a natural language narrative with a math question. The MWP solver is required to generate a solution equation, which can be calculated to get the numerical answer, by understanding the contextual problem description. 
    
In teaching, students are encouraged to recognize that mathematics is really about patterns and not merely about numbers~\citep{NAP1199}.
Mathematically excellent students explore patterns, not just memorize procedures~\citep{LearningToThink}.
Recently, \citet{nlp-able-solve-mwp} mention that existing MWP models~\citep{gts-seq2tree, g2t-graph2tree} rely on shallow heuristics to generate equations. These models can predict solutions well even if leaving only narratives without questions, which suggests that neural networks learn to solve MWPs by memorizing the lexical input like rote learning. Thus, existing models get stuck in memorize procedures.
We look at this issue and hypothesize it is because they focus on text understanding or equation generation for one problem. The same quantitative relationship corresponds to many problems of different themes and scenarios, but previous methods overlook the outlining and distinction of MWP patterns.



In this work, we first investigate how a neural network understands MWP patterns only from semantics. We adopt the widely used encoder-decoder model structure~\citep{gru}.
BERT~\citep{bert} is employed as the semantic encoder, and a tree decoder~\citep{gts-seq2tree} is adopted to generate equations. We probe the problem representations in BERT.
The visualization by T-SNE~\citep{tsne} in Figure~\ref{fig:intro-tsne} shows that, through the semantic encoder, most representations of problems with the same prototype equation are pulled closer, even if their narratives are semantically different. 
We also analyze the representations in different BERT layers, and the results show the lexical semantics mainly affects the problem-solving in lower layers.
Besides, for each prototype equation, those problem representations far away from its center representation tend to produce incorrect solutions.

Inspired by it, we propose a contrastive learning approach that seeks similar prototypes to support model to better understand patterns and perceive the divergence of patterns. 
When collecting contrastive examples, we follow~\citet{gts-seq2tree} to convert the prototype equation to a tree. Given an equation tree, the positive examples are retrieved if their trees or subtrees have the same structure, and the negative examples are collected from the rest in terms of the operator types and the size of the tree.
The solving model is first jointly optimized by an equation generation loss and a contrastive learning loss on the collected examples, and then, is further trained on the original dataset. While the generation loss empowers the model to memorize procedures from the semantics, the contrastive learning loss brings similar patterns closer and disperses the different patterns apart.
    
We conduct experiments on the Chinese dataset Math23k~\citep{math23k} and the English dataset MathQA~\citep{mathqa} in monolingual and multilingual settings. To support constructing multilingual contrastive examples, we follow~\citet{multilingual-mwp} to adapt MathQA as the counterpart of Math23k.
Experimental results show that our method achieves consistent gains in monolingual and multilingual settings.
In particular, our method allows the model to improve the performance in one language using data in another language, which suggests that MWP patterns are language-independent.
Furthermore, we verify that, through our contrastive learning, the representations that previously generate wrong solutions get closer to their centers, and several problems are solved well.

To summarize, the contributions of this paper include: 
i) An analysis of the MWP model showing that the semantic encoder understands lexical semantics in lower layers and gathers the prototype equations in higher layers.
ii) A contrastive learning approach helping the model to better understand MWP patterns and perceive the divergence of patterns.
iii) Applications in the multilingual setting suggesting that we can further improve the model performance using data in different languages.

\section{Related Work}

\begin{figure*}[t]
    \centering
    \includegraphics[width=15.5cm]{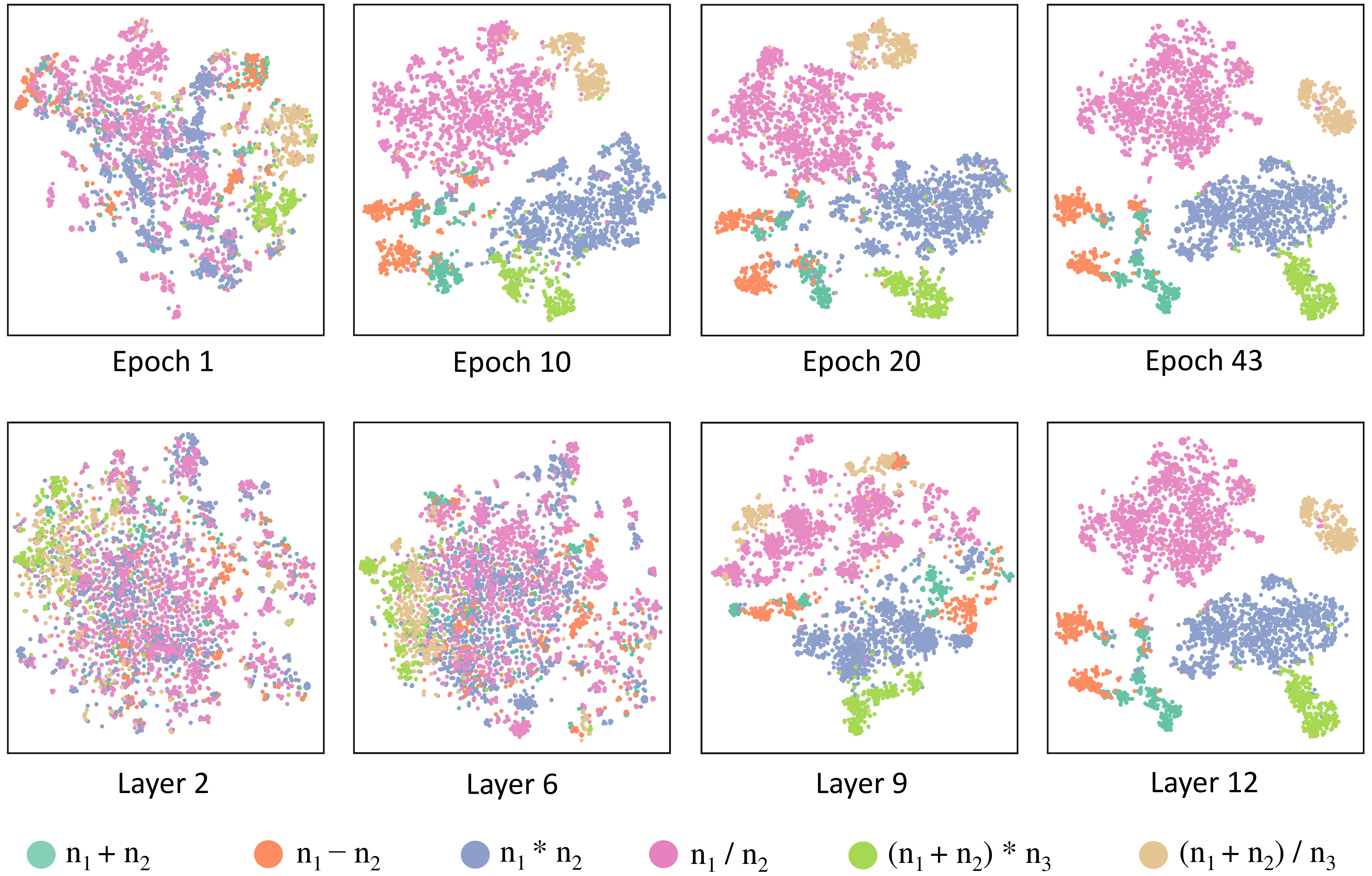}
    \caption{The T-SNE visualization of problem representations in different epochs and different layers. Different colors represent different prototype equations. The model achieves the highest accuracy at the training epoch 43. }
    \label{fig:s3-tsne}
\end{figure*}

\subsection{Math Word Problem Solving}
    Given a natural language narrative with a mathematical question, the task is to generate a solution equation to answer the question. The methods can be divided into four categories: rule-based methods~\citep{fletcher1985understanding, bakman2007robust}, statistical machine learning methods~\citep{kushman2014learning, hosseini2014learning}, semantic parsing methods~\citep{shi2015automatically, koncel2015parsing} and deep learning methods~\citep{math23k, danqing-mwp18, huang-etal-2018-using, gts-seq2tree, g2t-graph2tree}.
    
    Deep learning methods have achieved significant improvement on MWP solving. 
    \citet{math23k} first attempt to use recurrent neural networks to build a seq2seq solving model. 
    \citet{gts-seq2tree} propose a tree-structured decoder to generate an equation tree. Syntactically correct equations can be generated through traversing the equation tree. \citet{g2t-graph2tree} apply graph convolutional networks to extract relationships of quantities in math problems.
    Recently, unsupervised pretraining of language models~\citep{bert,xlnet} has provided informative contextual representations for text understanding, and fine-tuning techniques~\citep{cui-etal-2019-fine,li-etal-2021-improving-bert} have brought further performance gains.
    Several works~\citep{kim2020point, multilingual-mwp, gpt-mwp} based on pretrained language models enhance the ability of problem understanding.
    

\subsection{Contrastive Learning}
    Contrastive learning is a method of representation learning, which is first designed by \citet{hadsell2006dimensionality}. By pulling semantically similar embeddings together and pushing semantic different ones apart, contrastive learning can provide more effective representations. In NLP, similar approaches have been explored in many fields.
    \citet{bose-etal-2018-adversarial} develop a sampler to find harder negative examples, which forces the model to learn better word and graph embeddings.
    \citet{yang2019reducing} use contrastive learning to reduce word omission errors in neural machine translation.
    \citet{electra} train a discriminative model on contrastive examples to obtain more informative language representations.
    \citet{gao2021simcse} advance the performance of sentence embeddings by using contrastive learning in supervised and unsupervised settings. 
    \citet{yu-etal-2021-fine} develop a contrastive self-training to help language model fine-tuning and label denoising in weak supervision.

    To the best of our knowledge, this is the first work to adopt contrastive learning to MWP solving. With the supervision of contrastive learning, we seek similar MWP patterns to pull them closer, and collect confusing patterns to push them apart.
    
\section{Semantic Encoder Gathers Prototypes}

    
    
    In this section, we explore how a neural network understands patterns from semantics. We adopt the encoder-decoder model structure to solve problems, and perform analyses on the problem representations. 
    The observation is that the semantic encoder understands lexical semantics at lower layers and gathers the prototype equations at higher layers.
    
    

\subsection{Experimental Setup}

\subsubsection{Datasets}

We perform analyses on two widely used datasets Math23k~\citep{math23k} and MathQA~\citep{mathqa}. The Math23k dataset is composed of 23k MWPs in elementary education, and the MathQA has 37k MWPs with multiple choices and equations.

\subsubsection{Model Architecture}
\paragraph{Semantic Encoder}

    The pre-trained language model BERT~\citep{bert} is employed as the semantic encoder. The unsupervised pretraining on large corpora renders the model to learn linguistic knowledge, which provides rich textual representations.
    

\paragraph{Equation Decoder}
    A tree decoder \citep{gts-seq2tree} is adopted to generate solution equations. 
    We use the BERT-encoded representation of \texttt{[CLS]} token to initialize the root node when decoding.
    Recursively, the decoder generates the embedding of each node, and predicts the probabilities of number and operator candidates. 
    
    For brevity, we denote our model as \textbf{BERT-TD}. The model takes the textual problem description as the input and is optimized by minimizing the negative log-likelihoods of node probabilities for predicting the ground-truth equation tree.

\begin{figure}[t]
    \centering
    \includegraphics[width=7.0cm]{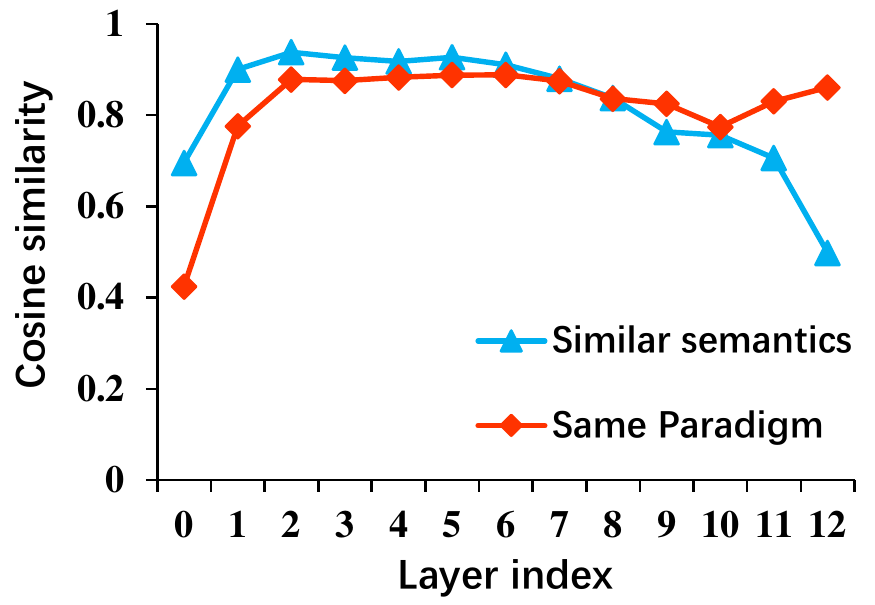}
    \caption{Similarities of problem representations in different BERT layers. The blue polyline corresponds to the semantically similar problems. The red polyline corresponds to problems with same prototype equation.}
    \label{fig:s3-sim}
\end{figure}

\subsection{Shifts of Problem Representation}
    \label{sec:shift-rep}
    To explore how the neural model learns MWP patterns during training, we first extract BERT-encoded representations of \texttt{[CLS]} token in different epochs and different layers. Then we perform the T-SNE visualization~\citep{tsne} shown in Figure~\ref{fig:s3-tsne}. The representations of different epochs are picked from the top layer of BERT, and the representations of different layers are picked from the best trained model.
    It can be seen that, as the training goes on, the representations with the same prototype equation are gathering.
    Besides, with the increase of the depth of encoder layers, the gathering tendency becomes more and more obvious.
    
    Intuitively, the prototype equation exhibits the essential relationship between the quantities in MWP. These results also verify that the patterns learned by the neural model are directly associated with the prototype equations.
    
\subsection{Semantics and Prototype Equation}
    From the visualizations, we can not see how semantics affects problem-solving. To this end, we collect 20 problem pairs with similar lexical semantics but exactly different prototypes, and 20 problem pairs with the same prototype but in different themes or scenarios. Not like taking the \texttt{[CLS]} representation in Section~\ref{sec:shift-rep}, we average the representations over all words in one problem. The cosine similarities of the averaged representations are calculated for these problem pairs in different BERT layers.
    
    The averaged similarities are shown in Figure~\ref{fig:s3-sim}. The semantically similar problems obtain higher values in lower layers but the similarity gradually decreases as the model deepens. Meanwhile, with the increase of the model depth, although in different semantics, the problems with the same prototype equation achieve higher similarity. 
    This demonstrates that lexical semantics affects problem-solving at lower layers, and the model further extracts prototypes from the semantics at higher layers.
    

\begin{figure}[t]
    \centering
    \includegraphics[width=6.85cm]{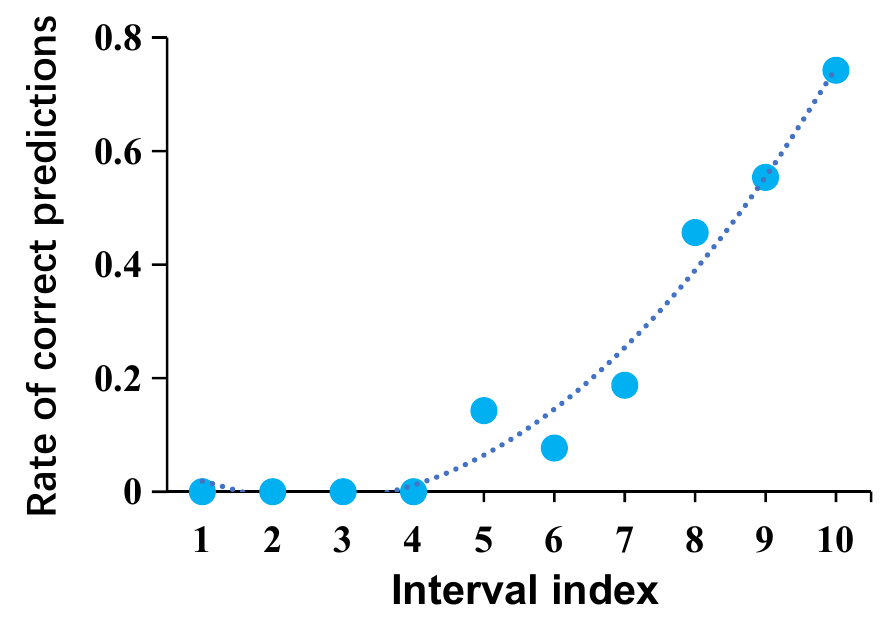}
    \caption{Model performance in each distance interval. The interval index $x$ indicates the cosine distances are in the interval $[0.1 \times (x-1) , 0.1 \times x)$. The dotted line is computed by polynomial least squares fitting.}
    \label{fig:s3-rate}
\end{figure}

\subsection{Clustering and Solving Ability}
\label{sec:clustering_and_solving}

With the above observation, we attempt to discover the relationship between prototype clustering and model performance.
For each prototype equation, we first average the representations of the corresponding problems to obtain its center point, and then calculate the cosine distances between representations and its centers. A higher cosine distance means the representation is closer to its center. We split the cosine distance into several intervals and compute the proportion of correct predictions for each interval. 
The results are shown in Figure~\ref{fig:s3-rate}, which suggests that the representations apart from centers tend to produce wrong solutions.




\begin{table}[t]
\centering
\small

\begin{tabular}{p{0.55\linewidth}c}
\toprule
Problem & Prototype Equation \\
\midrule
Larry starts with $n_1$ cards. $n_2$ are eaten by a hippopotamus. How many cards does Larry end with? & \textcolor{red}{$n_1-n_2$}\\
\midrule
Frank made $n_1$ dollars mowing lawns over the summer. If he spent $n_2$ dollars buying new mower blades, how many $n_3$ dollar games could he buy with the money he had left? & $\textcolor{red}{(n_1-n_2)}/n_3$ \\
\bottomrule
\end{tabular}

\caption{Math word problems with the same quantitative relationship, i.e. the subtraction of numerics $n_1$ and $n_2$. The same prototype equations are in red color.}
\label{tab:intro-cases}
\end{table}

\section{Contrastive Learning}

In this section, we propose a contrastive learning approach to help the model to perceive the divergence of MWP patterns.
One drawback of existing deep learning methods is that they overlook the outlining and distinction of MWP patterns. In contrast, we seek similar prototype equations from various problems to support model to understand patterns, and collect easily confused patterns for model to distinguish. 

\subsection{Data Collection}

    We construct contrastive MWP triples $(p, p^+, p^-)$ containing a basic problem $p$ and its positive and negative examples $\{p^+,p^-\}$.
    
\begin{figure}[t]
    \centering
      \includegraphics[width=7.5cm]{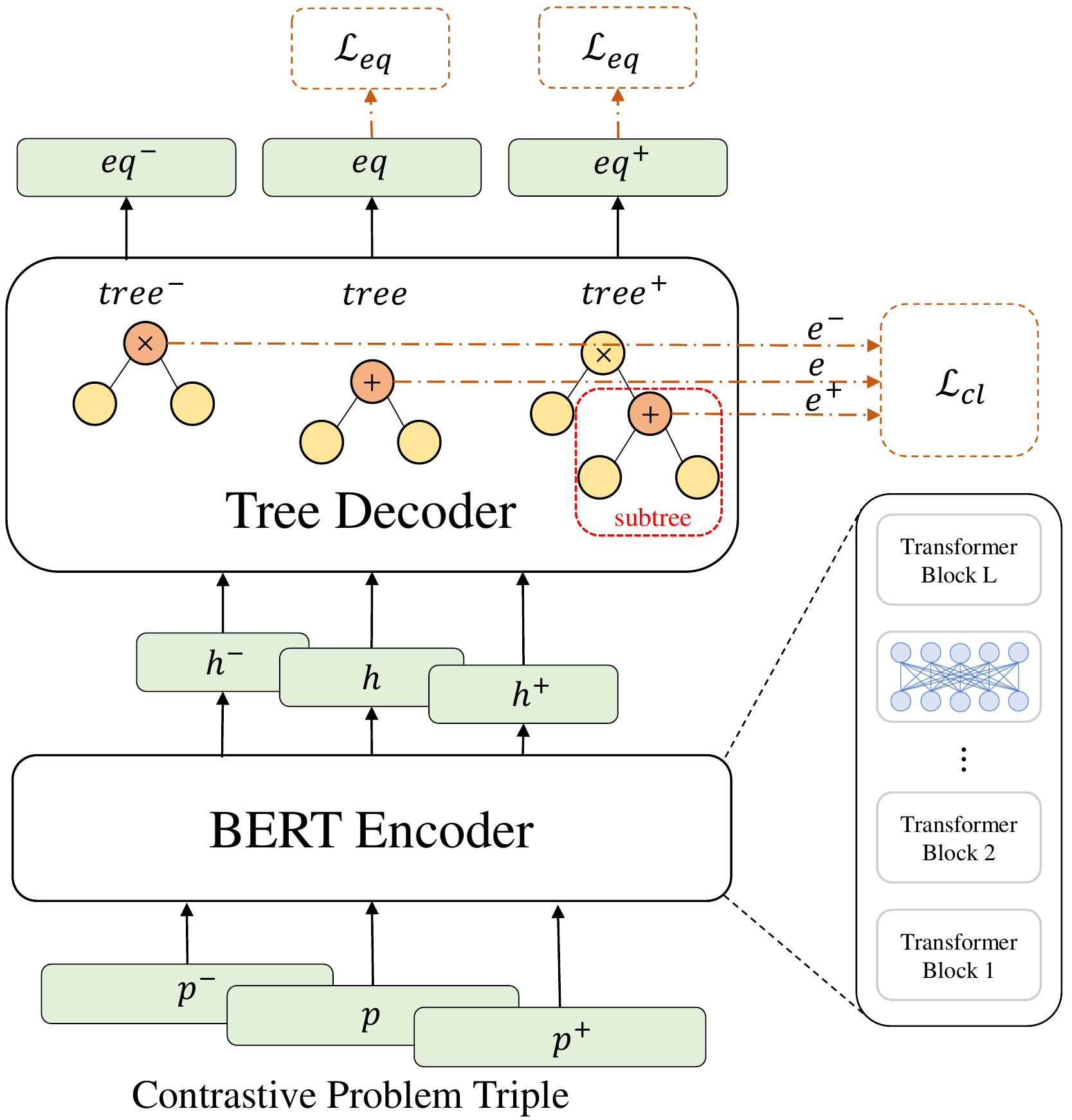}
    \caption{An overview of our model.}
    \label{fig:pipeline}
\end{figure}

    \paragraph{Positive Example}
    One direct way is to collect problems whose prototype equation is completely the same as the given problem $p$. However, the same quantitative relationship in $p$ also exists in other problems. As shown in Table~\ref{tab:intro-cases}, for the second problem, before answering "How many games could he buy?", another hidden question is "How much money does he have?" whose solving equation is in the same prototype as the first problem. Thus, we parse the prototype equation to tree structure by following~\citet{gts-seq2tree} and consider its sub-equations and subtrees. 
    The problem $p^+$ is taken as a positive example if its tree or subtree has the same structure as $p$, such as "$tree$" and the subtree of "$tree^+$" in Figure~\ref{fig:pipeline}. 

    \paragraph{Negative Example}
    \citet{bose-etal-2018-adversarial} and \citet{kalantidis2020hard} stress the importance of hard negative examples in contrastive learning. If we choose $p^-$ whose prototype is totally different from $p$, the original MWP model can easily distinguish them apart. Thus, in this work, the problem $p^-$ is chosen as a hard negative example if its tree has the same number of nodes but different operator node types, such as "$tree$" and "$tree^-$" in Figure~\ref{fig:pipeline}. With the training on hard negative examples, our model can distinguish more subtle differences from various prototypes, and further grasp the inner pattern of MWP.

    

\subsection{Training Procedure}
\label{sec:cl}

We train the model on our contrastive problem triples. As shown in Figure~\ref{fig:pipeline}, the problems are first encoded by BERT, and then the tree decoder predicts the nodes of the equation tree. 

During contrastive learning, the triple $z = (p, p^+, p^-)$ are input to the model together to predict equation trees. Owing to the decoding manner of~\citet{gts-seq2tree}, each node embedding represents the whole subtree information rooted in it.
The root node embeddings of the problem $p$ and its negative problem $p^-$ are picked for model to distinguish. 
For its positive problem $p^+$, we find the root node of the tree or subtree containing the same structure as $p$, and pull its embedding closer to that of $p$. For brevity, we denote these node embeddings as $(e, e^+, e^-)$ and the contrastive learning loss becomes:
\begin{equation}
\begin{split}
    \mathcal{L}_{cl} =
      \sum_{z} \max(0, \eta + sim(e, e^-) \\
      - sim(e, e^+))
    ,
\end{split}
\end{equation}
where $sim(\cdot)$ is the cosine similarity, and the $\eta$ is a margin hyper-parameter.

The basics of a MWP solving model is to generate a solution equation to answer the math question. We transform the target equation $y$ into Polish notation as $[y_1, y_2, ..., y_m]$, where $m$ is the equation length. The tree decoder generates $k$-node token $y_k$ recursively, and the loss of generating equation is computed as:
\begin{equation}
    \mathcal{P}(y|p) = \prod_{k=1}^m \mathcal{P}(y_k|p)
\end{equation}
\begin{equation}
    \mathcal{L}_{eq} = \sum_{p} -\log{\mathcal{P}(y|p)}
\end{equation}

The final training objective is to minimize the equation loss and contrastive loss as follows:
\begin{equation}
    \mathcal{L} = \mathcal{L}_{eq} + \alpha \cdot \mathcal{L}_{cl}
\end{equation}
where $\alpha$ is a hyper-parameter that represents the importance of the contrastive learning.

However, not all problems have positive examples, such as those problems whose solution is one value without any operator. With this in mind, we develop the two-stage training strategy. The MWP solver is first trained on our contrastive triples at stage I, and then further trained on the original dataset at stage II.

\begin{table}[t]
\centering

\begin{tabular}{lrrr}
\toprule
Dataset & \#Train & \#Dev & \#Test \\
\midrule
Math23k & 21,162 & 1,000 & 1,000 \\
\midrule
MathQA & 29,837 & 4,475 & 2,985 \\
MathQA$^\dagger$ & 23,703 & 3,540 & 2,410 \\

\bottomrule

\end{tabular}

\caption{Statistics of the used datasets. The "MathQA$^\dagger$" is the adapted MathQA dataset by following~\citet{multilingual-mwp}.}
\label{tab:data-stat}
\end{table}

\begin{table*}[t]
\centering
\begin{tabular}{lcccc}
\toprule
~ & \multicolumn{2}{c}{Math23k} & \multicolumn{2}{c}{MathQA$^\dagger$} \\
Models & Acc (eq) & Acc (ans) & Acc (eq) & Acc (ans) \\
\midrule
\multicolumn{5}{l}{\emph{Monolingual Setting}} \\
\quad GroupAttention~\citep{group-att} & - & 69.5 & ~63.3$^*$ & ~70.4$^*$ \\
\quad GTS~\citep{gts-seq2tree}            & - & 75.6 & ~68.9$^*$ & ~71.3$^*$ \\
\quad Graph2Tree~\citep{g2t-graph2tree}     & - & 77.4 & ~70.0$^*$ & ~72.0$^*$ \\
\quad BERT-TD w/o CL               & 71.2 & 82.4 & 73.5 & 75.1 \\
\quad BERT-TD w CL  & \textbf{71.8} & \textbf{83.2} & \textbf{74.4} & \textbf{76.3} \\
\midrule
\multicolumn{5}{l}{\emph{Multilingual Setting}} \\
\quad mBERT-TD w/o CL               & 67.8 & 80.5 & 72.0 & 73.5 \\
\quad mBERT-TD w CL  & \textbf{70.9} & \textbf{83.9} & \textbf{74.2} & \textbf{76.3} \\

\bottomrule

\end{tabular}

\caption{Main results on Math23k and the adapted MathQA test sets. "Acc(eq)" is the equation accuracy and "Acc(ans)" is the answer accuracy. "$*$" means our reimplementation based on released codes. "CL" is short for the contrastive learning. "mBERT" is short for the multilingual BERT.}
\label{tab:mono-res}
\end{table*}

\section{Experiments}
We evaluate our method on two widely used datasets~\citep{math23k,mathqa}, and demonstrate its effectiveness in monolingual and multilingual settings.
\subsection{Configuration}
\paragraph{Data and Metrics}
We collect problems from the Chinese dataset Math23k~\citep{math23k} and the English dataset MathQA~\citep{mathqa}. As the formula formats of the two datasets are different, we follow~\citet{multilingual-mwp} to adapt MathQA as a counterpart of Math23k.
Table~\ref{tab:data-stat} shows data statistics. 
We report the accuracy of equation generation, namely as "Acc (eq)", that the problem is solved well if the generated equation is equal to the annotated formula. Considering several equations satisfy the problem solution, we report the accuracy of answer value, namely as "Acc (ans)", to see whether the value calculated by the generated equation is equal to the target value.

\paragraph{Implementation}

We conduct our contrastive learning in the monolingual and multilingual perspectives. In the monolingual setting, we construct contrastive triples inside each dataset. In the multilingual setting, for each problem, the positive and negative examples are from different sources. Specifically, given a Chinese MWP in Math23k, we collect positive examples from MathQA and negative examples from Math23k. 
We adopt BERT-base~\citep{bert} as the problem encoder, and follow~\citet{gts-seq2tree} to build the tree-decoder for solution generation. The hidden size of the decoder is set to 768. Multilingual BERT is used in the multilingual setting.
The max input length is set to 120 and the max output length is set to 45.
The loss margin $\eta$ is set to 0.2. The weight $\alpha$ of contrastive learning loss is set to 5. 
We use AdamW~\citep{adamw} as our optimizer, and perform grid search over the sets of the learning rate as \{5e-5, 1e-4\} and the number of epochs as \{30, 50\} for each training stage.
The batch size is fixed to 16 to reduce the search space, and we evaluate models for every epoch. We use the dropout of 0.5 to prevent over-fitting and perform a 3-beam search for better generations.

\subsection{Baselines}
To verify the effectiveness of the proposed method, we directly train our model on original datasets without contrastive learning. In particular, the multilingual baseline model is trained by mixing Math23k and the adapted MathQA.
In addition to comparing with BERT, we also investigate the following approaches:

\textbf{GroupAttention}\footnote{\url{https://github.com/lijierui/group-attention}}~\citep{group-att} develop an attention mechanism to capture the quantity-related and question-related information.

\textbf{GTS}\footnote{\url{https://github.com/ShichaoSun/math_seq2tree}}~\citep{gts-seq2tree} generate equation trees through a tree structure decoder in a goal-driven mannner.

\textbf{Graph2Tree}\footnote{\url{https://github.com/2003pro/Graph2Tree}}~\citep{g2t-graph2tree} design a graph-based encoder for representing the relationships and order information among the quantities. 

\begin{table}[t]
\centering
\small
\begin{tabular}{l|cc|cc}
\toprule
& Pos. & Neg. & Math23k & MathQA$^\dagger$ \\
\midrule
Baseline & - & - & 80.5 & 73.5 \\
\midrule
\multirow{3}{*}{CL} & Same & Ours & 82.3 & 75.5 \\
& Ours & Rand & 82.3 & 75.8 \\
& Ours & Ours & \textbf{83.9} & \textbf{76.3} \\
\bottomrule

\end{tabular}

\caption{Results (answer accuracy) of different strategies collecting examples. "Pos." and "Neg." are corresponding to positive and negative examples. "Same" indicates the positive examples have exactly the same prototype equations. "Rand" indicates the negative examples are randomly selected from the rest.}
\label{tab:data-collect}
\end{table}

\subsection{Main Results}

Experimental results are shown in Table~\ref{tab:mono-res}. Training the MWP solver with our proposed contrastive learning outperforms the baseline models on all datasets.  

\paragraph{Monolingual Results} 
Compared to previous methods, the pretrained linguistic knowledge in BERT can help the MWP solver improve performance greatly. With our proposed contrastive learning method, our model achieves consistent gains on Math23k and the adapted MathQA. 
This suggests that seeking patterns with supervision benefits the model to solve MWPs.

\paragraph{Multilingual Results}
We adapt our model to the multilingual setting by using multilingual BERT and mixing two train sets. The contrastive learning improves Math23k answer accuracy to 83.9 (3.4 absolute improvements) and MathQA answer accuracy to 76.3 (2.8 absolute improvements), which are competitive with the monolingual results.
This demonstrates that the model can learn similar patterns in different languages.


\begin{table}[t]
\centering
\small
\begin{tabular}{l|ccccc}
\toprule
Margin~$\eta$ & 0.05 & 0.1 & 0.15 & 0.2 & 0.3 \\
\midrule
Math23k & 82.6 & 83.7 & 83.4 & \textbf{83.9} & 81.8 \\
MathQA$^\dagger$ & 76.1 & 76.2 & 76.1 & \textbf{76.3} & 76.0 \\
\bottomrule

\end{tabular}

\caption{Results (answer accuracy) of using different loss margin $\eta$ in the multilingual setting.}
\label{tab:margin}
\end{table}

\begin{table}[t]
\centering
\small
\begin{tabular}{ll|cc}

\toprule
& & Acc (eq) & Acc (ans) \\
\midrule
Baseline &  & 71.2 & 82.4 \\
\midrule
\multirow{2}{*}{CL ($\alpha=1$)} & Stage I & 70.1 & 81.5 \\
& Stage II & 70.5 & 83.0 \\
\midrule
\multirow{2}{*}{CL ($\alpha=5$)} & Stage I & 70.6 & 82.5 \\
& Stage II & \textbf{71.8} & \textbf{83.2} \\
\bottomrule

\end{tabular}

\caption{Results of using different loss weight $\alpha$ on Math23k in the monolingual setting. Two-stage results are reported.}
\label{tab:analysis-stage}
\end{table}

\subsection{Analysis}
We conduct ablations to better understand the contributions of different components in our contrastive learning method.

\subsubsection{Effects of Data Collection}

The contrastive examples consist of positive examples with similar patterns and negative examples with exactly different patterns.
In this work, we investigate different strategies of collecting positive and negative examples. As well as our strategy, we attempt to collect MWPs containing the same prototype equation to be the positive examples, and randomly select negative examples from the rest.

Table~\ref{tab:data-collect} shows that our strategy achieves better performance on all datasets. 
In addition to the problems with the same prototype equations, our collected examples include more problems having the same equation subtree structures. It can be seen that the model can benefit from these examples.
For the negative examples, we take the problems with the same number of operators but different operator types. If performing random selection, the model performance drops, which suggests that our collected examples can support the model to disperse the different patterns.
No matter which strategy we use, compared to the baseline without contrastive learning, our method advances MWP solving and gives one way to improve the performance by using data in different languages.

\begin{figure}[t]
    \centering
    \includegraphics[width=7.4cm]{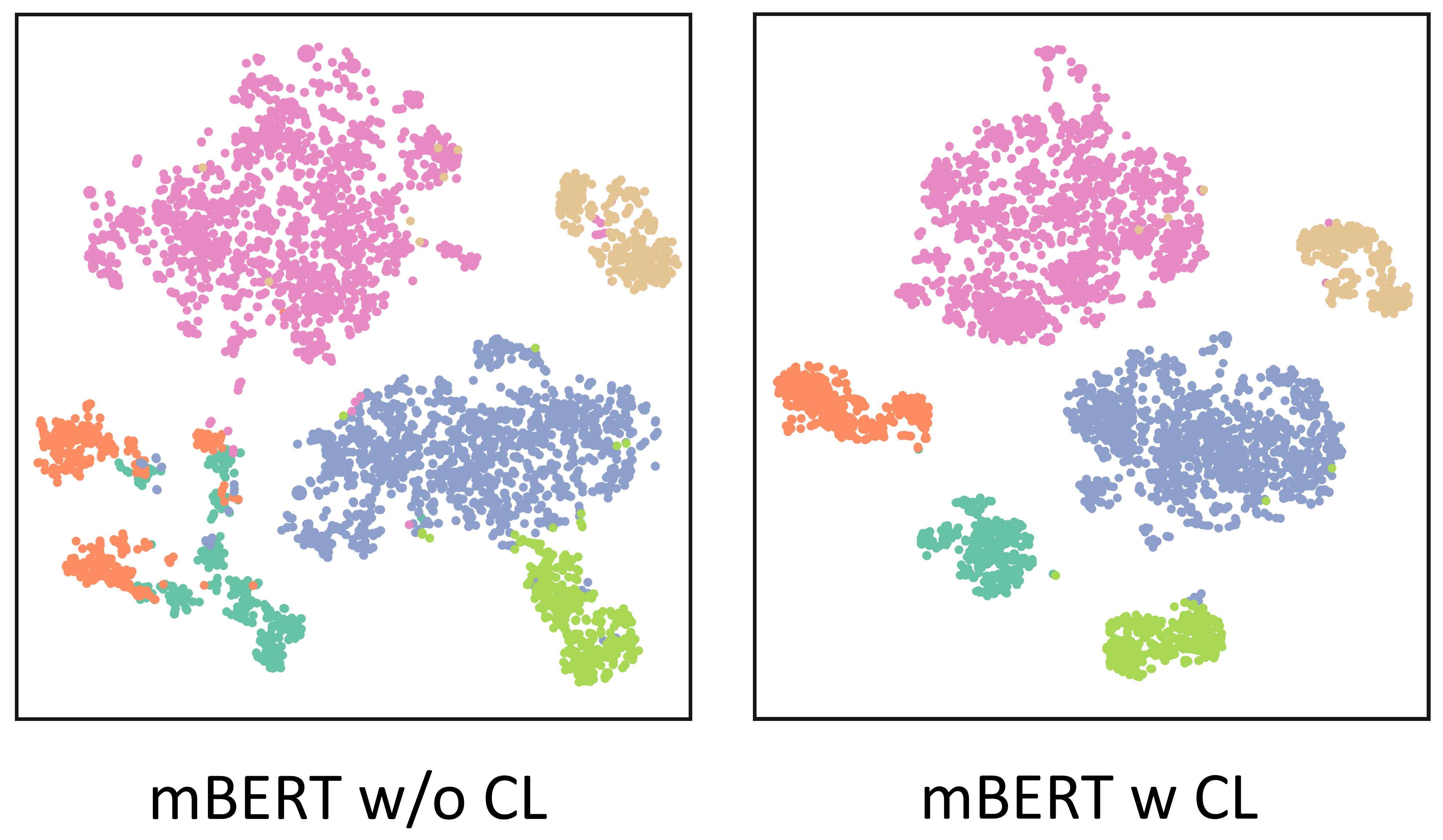}
    \caption{T-SNE visualization of the problem representation with and without our contrastive learning. 
    }
    \label{fig:analysis-tsne}
\end{figure}

\begin{figure}[t]
    \centering
    \includegraphics[width=7.4cm]{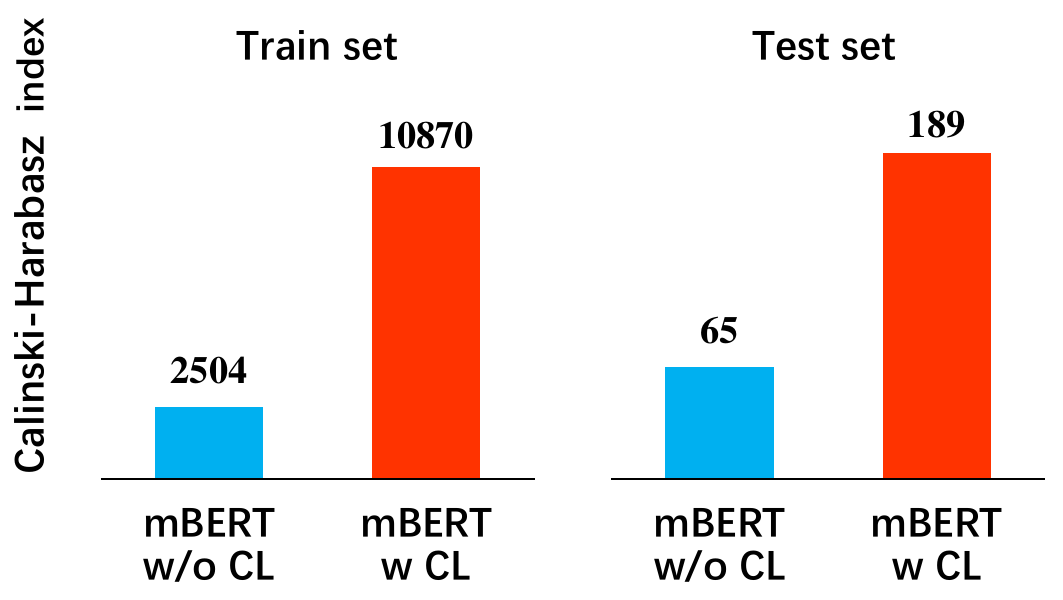}
    \caption{Calinski-Harabasz index on the train/test set with and without our contrastive learning.}
    \label{fig:analysis-ch-index}
\end{figure}

\subsubsection{Effects of Hyperparameters}

We train the "mBERT-TD" model with several loss margins (0.05, 0.1, 0.15, 0.2 and 0.3) to disperse the different patterns.   
As shown in Table~\ref{tab:margin}, the margin 0.2 can help the model achieve the best performance but lower margins 0.1 and 0.15 also perform well.

As introduced in Section~\ref{sec:cl}, we train our model in two stages and the loss weight $\alpha$ represents the importance of the contrastive learning. Table~\ref{tab:analysis-stage} shows the results of using different weights in each stage. It can be seen that the higher weight achieves better performance, and at stage II, training on all examples further improves the performance.





\subsubsection{Visualization and Statistics}

\begin{figure}[t]
    \centering
    \includegraphics[width=7.0cm]{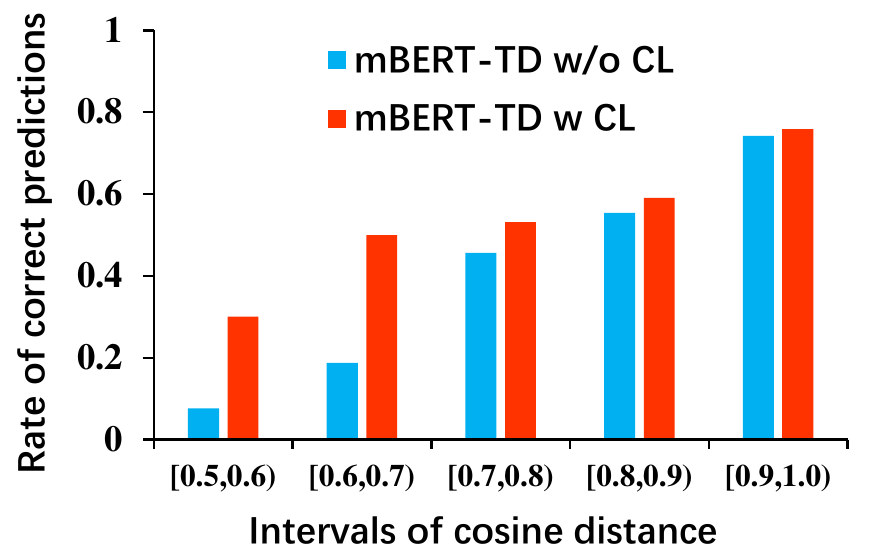}
    \caption{Equation accuracy in each distance interval with and without our contrastive learning.}
    \label{fig:analysis-rate}
\end{figure}

\begin{table}[t]
\centering
\small

\begin{tabular}{p{0.9\linewidth}}
\toprule
\textbf{Input:} A boatman selling a boat along river flow. If he sell boat in steal water at 3 m/sec and flow of river is 2 m/sec, how much time he will take to sell 100 m. \\
\textbf{Output (w/o CL):} 100 / (3 / 2) \\
\textbf{Output (w CL):} 100 / (3 + 2) \\
\midrule
\textbf{Input:} A pipe can fill the tank in 30 minutes and pipe b can empty the tank in 90 minutes. How long it will take to fill the tank if both pipes are operating together? \\
\textbf{Output (w/o CL):} 1 / ((1 / 30) + (1 / 90)) \\
\textbf{Output (w CL):}   1 / ((1 / 30) - (1 / 90)) \\
\midrule
\textbf{Input:} If 20 liters of chemical x are added to 80 liters of a mixture that is 25\% chemical x and 75\% chemical y, then what percentage of the resulting mixture is chemical x? \\
\textbf{Output (w/o CL):} 1 + ((25 / 100) * 5) \\
\textbf{Output (w CL):} 20 + ((25 / 100) * 80) \\
\bottomrule
\end{tabular}

\caption{Examples of the problem input and equation output of MWP solvers.}
\label{tab:case-study}
\end{table}

We perform the T-SNE visualization shown in Figure~\ref{fig:analysis-tsne}. The problem representations with the same prototype equation are more gathered through our contrastive learning. To measure this variation, we calculate the Calinski-Harabasz index~\citep{ch-index}.
Figure~\ref{fig:analysis-ch-index} shows that our method supports the model to gain higher clustering scores.

The above results illustrate that, for each prototype equation, the representations are pulled closer to its centers. We re-compute the proportion of correct predictions as described in Section~\ref{sec:clustering_and_solving}.
The results are shown in Figure~\ref{fig:analysis-rate}. 
We observe the accuracy increases in most intervals, which also verifies the effectiveness of contrastive learning. In particular, our model also performs well in lower intervals such as [0.6,0.7) and [0.7,0.8), which indicates those problems a little far away from their centers are not easily confused with other problems of different patterns, and our model disperses different patterns apart indeed.

Besides, we show few examples in Table~\ref{tab:case-study}. It can be seen that the contrastive learning method helps the model capture the quantitative relationships exactly.

\section{Conclusion}
In this paper, we find the neural network generates incorrect solutions due to the non-distinction of MWP patterns. To this end, we propose a contrastive learning approach to support the model to perceive divergence of patterns.
We seek similar patterns in terms of the equation tree structure and collect easily confused patterns for our model to distinguish. Our method outperforms previous baselines on Math23k and MathQA in monolingual and multilingual settings.

\bibliographystyle{acl_natbib}
\bibliography{main}

\begin{thebibliography}{34}
\expandafter\ifx\csname natexlab\endcsname\relax\def\natexlab#1{#1}\fi

\bibitem[{Amini et~al.(2019)Amini, Gabriel, Lin, Koncel-Kedziorski, Choi, and
  Hajishirzi}]{mathqa}
Aida Amini, Saadia Gabriel, Shanchuan Lin, Rik Koncel-Kedziorski, Yejin Choi,
  and Hannaneh Hajishirzi. 2019.
\newblock \href {https://doi.org/10.18653/v1/N19-1245} {{M}ath{QA}: Towards
  interpretable math word problem solving with operation-based formalisms}.
\newblock In \emph{Proceedings of the 2019 Conference of the North {A}merican
  Chapter of the Association for Computational Linguistics: Human Language
  Technologies, Volume 1 (Long and Short Papers)}, pages 2357--2367,
  Minneapolis, Minnesota. Association for Computational Linguistics.

\bibitem[{Bakman(2007)}]{bakman2007robust}
Yefim Bakman. 2007.
\newblock Robust understanding of word problems with extraneous information.
\newblock \emph{arXiv preprint math/0701393}.

\bibitem[{Bose et~al.(2018)Bose, Ling, and Cao}]{bose-etal-2018-adversarial}
Avishek~Joey Bose, Huan Ling, and Yanshuai Cao. 2018.
\newblock \href {https://doi.org/10.18653/v1/P18-1094} {Adversarial contrastive
  estimation}.
\newblock In \emph{Proceedings of the 56th Annual Meeting of the Association
  for Computational Linguistics (Volume 1: Long Papers)}, pages 1021--1032,
  Melbourne, Australia. Association for Computational Linguistics.

\bibitem[{Caliński and Harabasz(1974)}]{ch-index}
T.~Caliński and J~Harabasz. 1974.
\newblock \href {https://doi.org/10.1080/03610927408827101} {A dendrite method
  for cluster analysis}.
\newblock \emph{Communications in Statistics}, 3(1):1--27.

\bibitem[{Cho et~al.(2014)Cho, van Merrienboer, G{\"{u}}l{\c{c}}ehre, Bahdanau,
  Bougares, Schwenk, and Bengio}]{gru}
Kyunghyun Cho, Bart van Merrienboer, {\c{C}}aglar G{\"{u}}l{\c{c}}ehre, Dzmitry
  Bahdanau, Fethi Bougares, Holger Schwenk, and Yoshua Bengio. 2014.
\newblock \href {https://doi.org/10.3115/v1/d14-1179} {Learning phrase
  representations using {RNN} encoder-decoder for statistical machine
  translation}.
\newblock In \emph{Proceedings of the 2014 Conference on Empirical Methods in
  Natural Language Processing, {EMNLP} 2014, October 25-29, 2014, Doha, Qatar,
  {A} meeting of SIGDAT, a Special Interest Group of the {ACL}}, pages
  1724--1734. {ACL}.

\bibitem[{Clark et~al.(2020)Clark, Luong, Le, and Manning}]{electra}
Kevin Clark, Minh{-}Thang Luong, Quoc~V. Le, and Christopher~D. Manning. 2020.
\newblock \href {http://arxiv.org/abs/2003.10555} {{ELECTRA:} pre-training text
  encoders as discriminators rather than generators}.
\newblock \emph{CoRR}, abs/2003.10555.

\bibitem[{Cobbe et~al.(2021)Cobbe, Kosaraju, Bavarian, Hilton, Nakano, Hesse,
  and Schulman}]{gpt-mwp}
Karl Cobbe, Vineet Kosaraju, Mohammad Bavarian, Jacob Hilton, Reiichiro Nakano,
  Christopher Hesse, and John Schulman. 2021.
\newblock \href {http://arxiv.org/abs/2110.14168} {Training verifiers to solve
  math word problems}.
\newblock \emph{CoRR}, abs/2110.14168.

\bibitem[{Council(1989)}]{NAP1199}
National~Research Council. 1989.
\newblock \href {https://doi.org/10.17226/1199} {\emph{Everybody Counts: A
  Report to the Nation on the Future of Mathematics Education}}.
\newblock The National Academies Press, Washington, DC.

\bibitem[{Cui et~al.(2019)Cui, Li, Chen, and Zhang}]{cui-etal-2019-fine}
Baiyun Cui, Yingming Li, Ming Chen, and Zhongfei Zhang. 2019.
\newblock \href {https://doi.org/10.18653/v1/D19-1361} {Fine-tune {BERT} with
  sparse self-attention mechanism}.
\newblock In \emph{Proceedings of the 2019 Conference on Empirical Methods in
  Natural Language Processing and the 9th International Joint Conference on
  Natural Language Processing (EMNLP-IJCNLP)}, pages 3548--3553, Hong Kong,
  China. Association for Computational Linguistics.

\bibitem[{Devlin et~al.(2019)Devlin, Chang, Lee, and Toutanova}]{bert}
Jacob Devlin, Ming{-}Wei Chang, Kenton Lee, and Kristina Toutanova. 2019.
\newblock \href {https://doi.org/10.18653/v1/n19-1423} {{BERT:} pre-training of
  deep bidirectional transformers for language understanding}.
\newblock In \emph{Proceedings of the 2019 Conference of the North American
  Chapter of the Association for Computational Linguistics: Human Language
  Technologies, {NAACL-HLT} 2019, Minneapolis, MN, USA, June 2-7, 2019, Volume
  1 (Long and Short Papers)}, pages 4171--4186. Association for Computational
  Linguistics.

\bibitem[{Fletcher(1985)}]{fletcher1985understanding}
Charles~R Fletcher. 1985.
\newblock Understanding and solving arithmetic word problems: A computer
  simulation.
\newblock \emph{Behavior Research Methods, Instruments, \& Computers},
  17(5):565--571.

\bibitem[{Gao et~al.(2021)Gao, Yao, and Chen}]{gao2021simcse}
Tianyu Gao, Xingcheng Yao, and Danqi Chen. 2021.
\newblock Simcse: Simple contrastive learning of sentence embeddings.
\newblock \emph{arXiv preprint arXiv:2104.08821}.

\bibitem[{Hadsell et~al.(2006)Hadsell, Chopra, and
  LeCun}]{hadsell2006dimensionality}
Raia Hadsell, Sumit Chopra, and Yann LeCun. 2006.
\newblock Dimensionality reduction by learning an invariant mapping.
\newblock In \emph{2006 IEEE Computer Society Conference on Computer Vision and
  Pattern Recognition (CVPR'06)}, volume~2, pages 1735--1742. IEEE.

\bibitem[{Hosseini et~al.(2014)Hosseini, Hajishirzi, Etzioni, and
  Kushman}]{hosseini2014learning}
Mohammad~Javad Hosseini, Hannaneh Hajishirzi, Oren Etzioni, and Nate Kushman.
  2014.
\newblock Learning to solve arithmetic word problems with verb categorization.
\newblock In \emph{Proceedings of the 2014 Conference on Empirical Methods in
  Natural Language Processing (EMNLP)}, pages 523--533.

\bibitem[{Huang et~al.(2018{\natexlab{a}})Huang, Liu, Lin, and
  Yin}]{danqing-mwp18}
Danqing Huang, Jing Liu, Chin-Yew Lin, and Jian Yin. 2018{\natexlab{a}}.
\newblock \href {https://aclanthology.org/C18-1018} {Neural math word problem
  solver with reinforcement learning}.
\newblock In \emph{Proceedings of the 27th International Conference on
  Computational Linguistics}, pages 213--223, Santa Fe, New Mexico, USA.
  Association for Computational Linguistics.

\bibitem[{Huang et~al.(2018{\natexlab{b}})Huang, Yao, Lin, Zhou, and
  Yin}]{huang-etal-2018-using}
Danqing Huang, Jin-Ge Yao, Chin-Yew Lin, Qingyu Zhou, and Jian Yin.
  2018{\natexlab{b}}.
\newblock \href {https://doi.org/10.18653/v1/P18-1039} {Using intermediate
  representations to solve math word problems}.
\newblock In \emph{Proceedings of the 56th Annual Meeting of the Association
  for Computational Linguistics (Volume 1: Long Papers)}, pages 419--428,
  Melbourne, Australia. Association for Computational Linguistics.

\bibitem[{Kalantidis et~al.(2020)Kalantidis, Sariyildiz, Pion, Weinzaepfel, and
  Larlus}]{kalantidis2020hard}
Yannis Kalantidis, Mert~Bulent Sariyildiz, Noe Pion, Philippe Weinzaepfel, and
  Diane Larlus. 2020.
\newblock Hard negative mixing for contrastive learning.
\newblock \emph{arXiv preprint arXiv:2010.01028}.

\bibitem[{Kim et~al.(2020)Kim, Ki, Lee, and Gweon}]{kim2020point}
Bugeun Kim, Kyung~Seo Ki, Donggeon Lee, and Gahgene Gweon. 2020.
\newblock Point to the expression: Solving algebraic word problems using the
  expression-pointer transformer model.
\newblock In \emph{Proceedings of the 2020 Conference on Empirical Methods in
  Natural Language Processing (EMNLP)}, pages 3768--3779.

\bibitem[{Koncel-Kedziorski et~al.(2015)Koncel-Kedziorski, Hajishirzi,
  Sabharwal, Etzioni, and Ang}]{koncel2015parsing}
Rik Koncel-Kedziorski, Hannaneh Hajishirzi, Ashish Sabharwal, Oren Etzioni, and
  Siena~Dumas Ang. 2015.
\newblock Parsing algebraic word problems into equations.
\newblock \emph{Transactions of the Association for Computational Linguistics},
  3:585--597.

\bibitem[{Kushman et~al.(2014)Kushman, Artzi, Zettlemoyer, and
  Barzilay}]{kushman2014learning}
Nate Kushman, Yoav Artzi, Luke Zettlemoyer, and Regina Barzilay. 2014.
\newblock Learning to automatically solve algebra word problems.
\newblock In \emph{Proceedings of the 52nd Annual Meeting of the Association
  for Computational Linguistics (Volume 1: Long Papers)}, pages 271--281.

\bibitem[{Li et~al.(2019)Li, Wang, Zhang, Wang, Dai, and Zhang}]{group-att}
Jierui Li, Lei Wang, Jipeng Zhang, Yan Wang, Bing~Tian Dai, and Dongxiang
  Zhang. 2019.
\newblock \href {https://doi.org/10.18653/v1/P19-1619} {Modeling intra-relation
  in math word problems with different functional multi-head attentions}.
\newblock In \emph{Proceedings of the 57th Annual Meeting of the Association
  for Computational Linguistics}, pages 6162--6167, Florence, Italy.
  Association for Computational Linguistics.

\bibitem[{Li et~al.(2021)Li, Zhou, Li, Xu, and
  Cao}]{li-etal-2021-improving-bert}
Zhongli Li, Qingyu Zhou, Chao Li, Ke~Xu, and Yunbo Cao. 2021.
\newblock \href {https://doi.org/10.18653/v1/2021.findings-acl.57} {Improving
  {BERT} with syntax-aware local attention}.
\newblock In \emph{Findings of the Association for Computational Linguistics:
  ACL-IJCNLP 2021}, pages 645--653, Online. Association for Computational
  Linguistics.

\bibitem[{Loshchilov and Hutter(2017)}]{adamw}
Ilya Loshchilov and Frank Hutter. 2017.
\newblock \href {http://arxiv.org/abs/1711.05101} {Fixing weight decay
  regularization in adam}.
\newblock \emph{CoRR}, abs/1711.05101.

\bibitem[{Patel et~al.(2021)Patel, Bhattamishra, and
  Goyal}]{nlp-able-solve-mwp}
Arkil Patel, Satwik Bhattamishra, and Navin Goyal. 2021.
\newblock \href {https://doi.org/10.18653/v1/2021.naacl-main.168} {Are {NLP}
  models really able to solve simple math word problems?}
\newblock In \emph{Proceedings of the 2021 Conference of the North American
  Chapter of the Association for Computational Linguistics: Human Language
  Technologies}, pages 2080--2094, Online. Association for Computational
  Linguistics.

\bibitem[{Schoenfeld(1992)}]{LearningToThink}
A.~Schoenfeld. 1992.
\newblock Learning to think mathematically: Problem solving, metacognition, and
  sense making in mathematics (reprint).
\newblock \emph{Journal of Education}, 196:1 -- 38.

\bibitem[{Shi et~al.(2015)Shi, Wang, Lin, Liu, and Rui}]{shi2015automatically}
Shuming Shi, Yuehui Wang, Chin-Yew Lin, Xiaojiang Liu, and Yong Rui. 2015.
\newblock Automatically solving number word problems by semantic parsing and
  reasoning.
\newblock In \emph{Proceedings of the 2015 Conference on Empirical Methods in
  Natural Language Processing}, pages 1132--1142.

\bibitem[{Tan et~al.(2021)Tan, Wang, Jiang, and Jiang}]{multilingual-mwp}
Minghuan Tan, Lei Wang, Lingxiao Jiang, and Jing Jiang. 2021.
\newblock \href {http://arxiv.org/abs/2105.08928} {Investigating math word
  problems using pretrained multilingual language models}.

\bibitem[{van~der Maaten and Hinton(2008)}]{tsne}
Laurens van~der Maaten and Geoffrey Hinton. 2008.
\newblock \href {http://www.jmlr.org/papers/v9/vandermaaten08a.html}
  {Visualizing data using {t-SNE}}.
\newblock \emph{Journal of Machine Learning Research}, 9:2579--2605.

\bibitem[{Wang et~al.(2017)Wang, Liu, and Shi}]{math23k}
Yan Wang, Xiaojiang Liu, and Shuming Shi. 2017.
\newblock \href {https://doi.org/10.18653/v1/D17-1088} {Deep neural solver for
  math word problems}.
\newblock In \emph{Proceedings of the 2017 Conference on Empirical Methods in
  Natural Language Processing}, pages 845--854, Copenhagen, Denmark.
  Association for Computational Linguistics.

\bibitem[{Xie and Sun(2019)}]{gts-seq2tree}
Zhipeng Xie and Shichao Sun. 2019.
\newblock \href {https://doi.org/10.24963/ijcai.2019/736} {A goal-driven
  tree-structured neural model for math word problems}.
\newblock In \emph{Proceedings of the Twenty-Eighth International Joint
  Conference on Artificial Intelligence, {IJCAI-19}}, pages 5299--5305.
  International Joint Conferences on Artificial Intelligence Organization.

\bibitem[{Yang et~al.(2019{\natexlab{a}})Yang, Dai, Yang, Carbonell,
  Salakhutdinov, and Le}]{xlnet}
Zhilin Yang, Zihang Dai, Yiming Yang, Jaime Carbonell, Ruslan Salakhutdinov,
  and Quoc~V Le. 2019{\natexlab{a}}.
\newblock {XLNet}: Generalized autoregressive pretraining for language
  understanding.
\newblock \emph{arXiv preprint arXiv:1906.08237}.

\bibitem[{Yang et~al.(2019{\natexlab{b}})Yang, Cheng, Liu, and
  Sun}]{yang2019reducing}
Zonghan Yang, Yong Cheng, Yang Liu, and Maosong Sun. 2019{\natexlab{b}}.
\newblock Reducing word omission errors in neural machine translation: A
  contrastive learning approach.
\newblock In \emph{Proceedings of the 57th Annual Meeting of the Association
  for Computational Linguistics}, pages 6191--6196.

\bibitem[{Yu et~al.(2021)Yu, Zuo, Jiang, Ren, Zhao, and
  Zhang}]{yu-etal-2021-fine}
Yue Yu, Simiao Zuo, Haoming Jiang, Wendi Ren, Tuo Zhao, and Chao Zhang. 2021.
\newblock \href {https://doi.org/10.18653/v1/2021.naacl-main.84} {Fine-tuning
  pre-trained language model with weak supervision: A contrastive-regularized
  self-training approach}.
\newblock In \emph{Proceedings of the 2021 Conference of the North American
  Chapter of the Association for Computational Linguistics: Human Language
  Technologies}, pages 1063--1077, Online. Association for Computational
  Linguistics.

\bibitem[{Zhang et~al.(2020)Zhang, Wang, Lee, Bin, Wang, Shao, and
  Lim}]{g2t-graph2tree}
Jipeng Zhang, Lei Wang, Roy Ka-Wei Lee, Yi~Bin, Yan Wang, Jie Shao, and Ee-Peng
  Lim. 2020.
\newblock \href {https://doi.org/10.18653/v1/2020.acl-main.362} {Graph-to-tree
  learning for solving math word problems}.
\newblock In \emph{Proceedings of the 58th Annual Meeting of the Association
  for Computational Linguistics}, pages 3928--3937, Online. Association for
  Computational Linguistics.

\end{thebibliography}


\end{document}